\documentclass[sigconf,nonacm]{acmart}

\AtBeginDocument{%
  }
\renewcommand\footnotetextcopyrightpermission[1]{}
\setcopyright{acmlicensed}
\copyrightyear{2018}
\acmYear{2018}
\acmDOI{XXXXXXX.XXXXXXX}

\acmConference[WWW '25]{Make sure to enter the correct
  conference title from your rights confirmation emai}{April 28--May 02,
  2025}{Sydney, Australia}
\acmISBN{978-1-4503-XXXX-X/18/06}

\usepackage{amsthm,amsmath}
\usepackage{mathrsfs}
\usepackage{longtable}
\usepackage{multirow} 
\usepackage{longtable}
\usepackage{booktabs}
\usepackage{xcolor}
\usepackage{graphicx} 
\definecolor{yellow}{RGB}{255,250,205}
\begin{document}

\title{AgentCTG: Harnessing Multi-Agent Collaboration for Fine-Grained Precise Control in Text Generation}

\author{Xinxu Zhou}
\email{zhouxinxu.zxx@alibaba-inc.com}
\affiliation{
  \institution{AMAP, Alibaba Group}
  \city{Beijing}
  \country{China}
}

\author{Jiaqi Bai}
\email{baijiaqi.bjq@alibaba-inc.com}
\affiliation{
  \institution{AMAP, Alibaba Group}
  \city{Beijing}
  \country{China}
}

\author{Zhenqi Sun}
\email{sunzhenqi.szq@alibaba-inc.com}
\affiliation{
  \institution{AMAP, Alibaba Group}
  \city{Beijing}
  \country{China}
}

\author{Fanxiang Zeng}
\email{fanxiang.zfx@alibaba-inc.com}
\affiliation{
  \institution{AMAP, Alibaba Group}
  \city{Beijing}
  \country{China}
}

\author{Yue Liu}
\email{yue.liu@alibaba-inc.com}
\affiliation{
  \institution{AMAP, Alibaba Group}
  \city{Beijing}
  \country{China}
}

\renewcommand{\shortauthors}{Zhou, et al.}
\begin{abstract}
Although significant progress has been made in many tasks within the field of Natural Language Processing (NLP), Controlled Text Generation (CTG) continues to face numerous challenges, particularly in achieving fine-grained conditional control over generation. Additionally, in real scenario and online applications, cost considerations, scalability, domain knowledge learning and more precise control are required, presenting more challenge for CTG. This paper introduces a novel and scalable framework, AgentCTG, which aims to enhance precise and complex control over the text generation by simulating the control and regulation mechanisms in multi-agent workflows. We explore various collaboration methods among different agents and introduce an auto-prompt module to further enhance the generation effectiveness. AgentCTG achieves state-of-the-art results on multiple public datasets. To validate its effectiveness in practical applications, we propose a new challenging Character-Driven Rewriting task, which aims to convert the original text into new text that conform to specific character profiles and simultaneously preserve the domain knowledge. When applied to online navigation with role-playing, our approach significantly enhances the driving experience through improved content delivery. By optimizing the generation of contextually relevant text, we enable a more immersive interaction within online communities, fostering greater personalization and user engagement.
\end{abstract}

%



\begin{teaserfigure}
  \includegraphics[width=\textwidth]{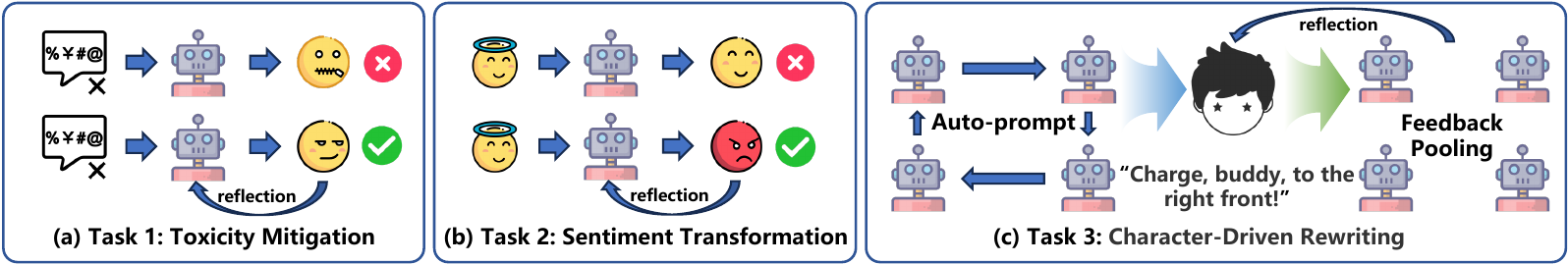}
  \caption{Illustration of different Controlled Text Generation (CTG) tasks. Traditionally, methods such as prompt engineering or retraining models to control the properties of text often fail, particularly in complex real-world tasks like the Character-Driven Rewriting task depicted in (c). To address this issue, we propose a noval multi-agent collaboration framework, where auto-prompt generation and reflection mechanisms are integrated, significantly enhancing the controllability of the generated text and making it more aligned with the desired attributes and objectives.}
  \label{fig:teaser}
\end{teaserfigure}

\maketitle

\section{Introduction}
Controlled text generation (CTG) \cite{zhang2023survey} refers to the practice of regulating and guiding the text produced according to specific constraints \cite{li2024pre} or objectives \cite{wang2024controllable}, ensuring that the generated text meets the intended requirements. This research area has garnered widespread attention in recent years \cite{ashok2024controllable}, mainly due to its potential across various applications, including sentiment control \cite{dathathriplug}, content production \cite{keskar2019ctrl}, story writing \cite{chen2024reflections}, and scriptwriting \cite{wu2024role, patel2024swag}. CTG methods not only enhance the quality, diversity, and personalization of online content but also ensures that the output is engaging for target audiences, thereby improving user experience and driving innovation in the digital content industry.

Traditionally, CTG methods employ deep neural network approaches to achieve basic control over the attributes of generated text, including adversarial learning \cite{hu2017toward, xu2019modeling, prabhumoye2020exploring} and diffusion learning \cite{li2022diffusion, he2023diffusionbert}. With the rapid development of large language models (LLMs) \cite{chang2024survey}, more research \cite{liang2024controlled, zhong2023air} increasingly focus on using smaller language models to influence the decoding process of LLMs. While the approaches provides a certain degree of control, its potential shortcomings have gradually become apparent: when smaller models dominate decision-making, they may obscure the original performance and potential of LLMs, limiting their effectiveness across various contexts and ultimately leading to a decline in the fluency of the generated text.

Currently, researches on LLMs instruction following \cite{lou2024large} predominantly focus on tasks involving simple instructions and short responses \cite{zhou2023controlled, yuan2024following}. 
However, in practical industrial applications, text generation faces more challenges of increasingly multifaceted, fine-grained control and thereby direct usage of LLMs can be unsatisfactory.
For example, general LLMs may struggle to accurately understand the eight directions in navigation scenarios, such as incorrectly equating "forward right" with "turn right". This phenomenon arises from the complexity and uniqueness of specialized domain.
In complex scenarios, LLMs may have difficulty accurately understanding the context. Particularly in cases requiring precise control or fine-grained information, the generated outputs may deviate from expectations. To produce engaging and practically valuable text, creators must continuously adjust drafts based on external feedback, adopting the user's perspective to establish a deep emotional connection. Therefore, prompt engineering may not be sufficient to ensure that LLMs effectively adhere to stringent instructions.

To explore the potential of LLMs in fine-grained CTG tasks and creative text generation tasks, we focus on tasks involving Toxicity Mitigation \cite{dathathriplug} and Sentiment Transformation \cite{pei2023preadd}, as illustrated in Figure \ref{fig:teaser}. To evaluate our model's adherence to complex instructions, we introduce the new Character-Driven Rewriting task. This task is designed to rewrite text within a new instruction framework that encompasses various dimensions, including character setting, content accuracy, and word count control. Through detailed and diversified instructions, we aim to comprehensively assess the model's performance in handling complex scenarios, ensuring that the generated text aligns with specific character traits while meeting accuracy and formatting requirements for practical applications. We design a fully automated multi-role framework called \textbf{AgentCTG}, aimed at enabling LLMs to emulate the creative processes of skilled experts. Specifically, we organize LLMs into different roles that participate in the human creative process, including writers and quality inspectors. Additionally, to generate higher-quality prompts that are easier for LLMs to follow, we introduce an additional auto-prompt generation progress. We emphasize the collaboration methods within and between different modules. Each step in our framework operates independently, allowing for easy human intervention at any stage. In summary, our contributions are as follows:
\begin{itemize}
    \item We introduce multi-agent collaboration in the field of CTG, providing new insights into the use of LLMs for text generation. In AgentCTG, we explore various collaborative mechanisms while incorporating an auto-prompt mechanism to enable LLMs to perform operations more effectively. And we propose a new CTG task, Character-Driven Rewriting, which presents more complex control conditions and lower fault tolerance, to demonstrate the practical application value of the model.
    \item AgentCTG achieves state-of-the-art results on both public and private datasets. Additionally, the atomized multi-agent framework provides our model with high generalization ability and scalability. We also develop the Human Review Preference Evaluation (HRPE) strategy to assess the model's performance in Character-Driven Rewriting task. This strategy offers more precise guidance for real-world applications, helping us optimize output quality and user experience during the text generation process.
    \item We release a new CTG dataset focused on the navigation instruction domain to facilitate researches in CTG. Users need only provide simple character setting descriptions as input; leveraging our fully automated process framework, our method can automatically handle complex tasks without requiring users to possess extensive knowledge of the navigation domain.
\end{itemize}

\section{Related Work}

\subsection{Controlled Text Generation}
Controlled Text Generation (CTG) is viewed as a supervised sequence-to-sequence problem under the encoding-decoding paradigm \cite{sutskever2014sequence}. Previous research focus on designing attribute mapping techniques for CTG that maximize joint likelihood through variational inference \cite{oussidi2018deep} and conditional sequence generation \cite{wu2023ar}. However, in practical applications, the controllability of the generated text may still be limited, making it difficult to meet specific user needs or preferences.

As interest in CTG methods based on pretrained models \cite{zhang2023survey} gradually increases, researchers propose various innovative solutions. Keskar et al. \cite{keskar2019ctrl} introduces CTRL, which integrates control mechanisms into the model architecture through retraining. Although this method achieves a higher level of controllability, it requires a large amount of high-quality data with attribute labels and incurs substantial training time costs, limiting its potential for widespread application. Some studies \cite{liang2024controlled, zhang2020dialogpt, lin2021adapter, zeldes2020technical} propose strategies using smaller models to influence the decoding process of LLMs. They train additional attribute models to affect the output logits distribution of LLMs through a plug-and-play approach, enabling attribute control of the generated text. However, these methods often face limitations in controlling a single attribute, and modifying the original model’s output can significantly undermine the fluency and contextual coherence of the generated text. The progress of instruction-based models is also noteworthy, such as FLAN \cite{wei2021finetuned} and InstructCTG \cite{ouyang2022training}. These methods make significant advances in zero-shot learning performance, achieving good results in simple attribute control. However, when facing complex real-world scenarios, instruction-based models may struggle to accurately understand and respond to complex requests. Additionally, many instruction-based models lack flexibility in addressing new types of tasks or inputs, making it difficult to quickly adapt to specific user requirements.
\begin{figure*}[h]
  \centering
  \includegraphics[width=\linewidth]{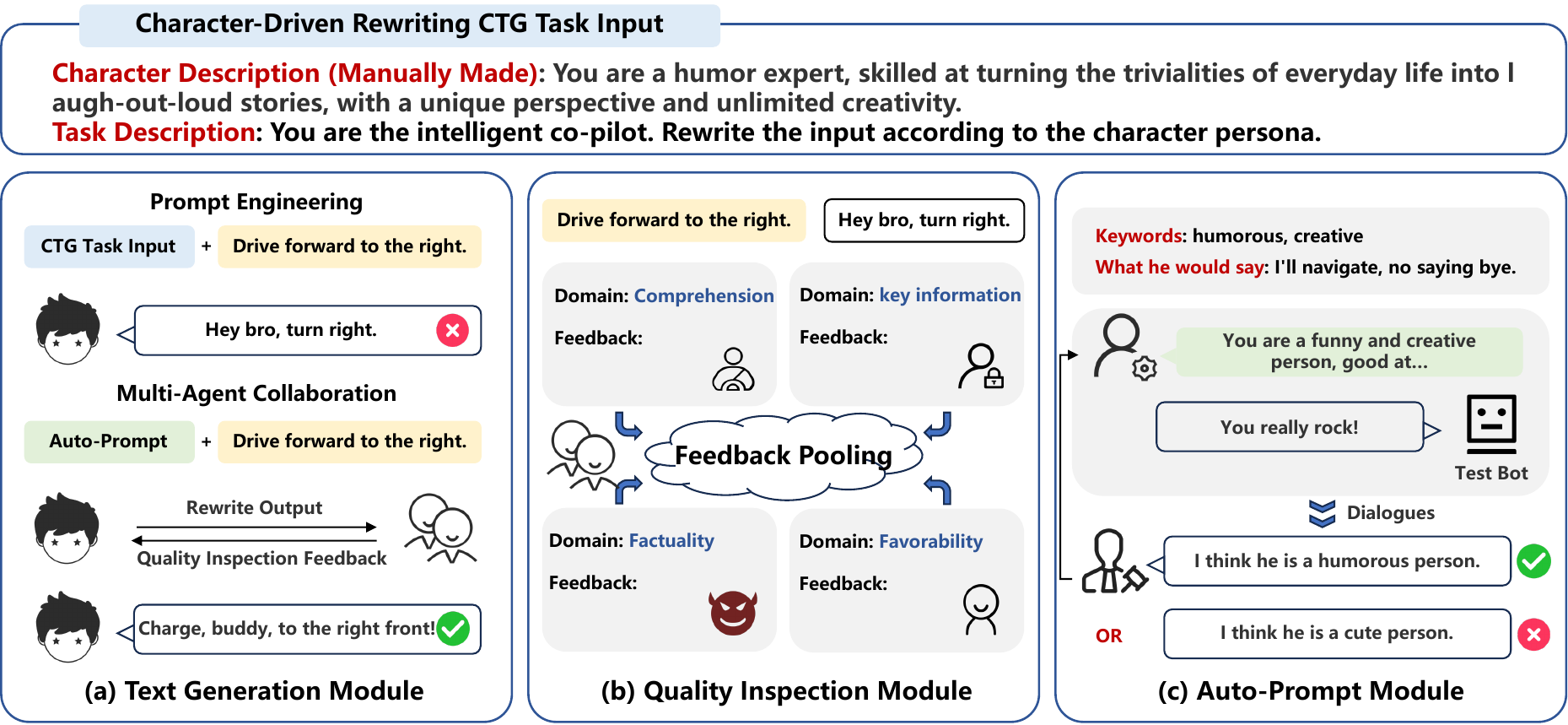}
  \caption{The overview of the AgentCTG. Taking the Character-Driven Rewriting task as an example, the CTG Task Input represents the original prompt provided by human. (a) denotes the Text Generation process, where the Decentralized Quality Inspection module is represented by (b), and the Free Performance-based Auto-Prompt module is represented by (c).}
  \label{fig:figure2}
  \vspace{-5pt}
\end{figure*}

\subsection{Multi-Agent Collaboration}
In the field of Artificial Intelligent agents, the primary responsibility of the agent is to perceive its environment \cite{park2023generative}, make rational decisions \cite{wang2024multimodal}, and take appropriate actions to respond to diverse and complex situations \cite{sheng2023learning}. The robust capabilities of LLMs align closely with these requirements, thereby accelerating the development of multi-agent systems across various application scenarios. Compared to single-agent systems, multi-agent systems can more effectively solve complex problems through collaboration among agents or by utilizing combinations of different agent roles to realistically simulate intricate real-world environments \cite{xi2023rise, guo2024large}. 

Researches have shown that multi-agent systems perform well in complex tasks such as literary translation \cite{wu2024perhaps}, medical diagnosis \cite{shi2024medadapter}, and script writing \cite{chen2024hollmwood}. Furthermore, researches employ multiple agents to simulate real-world environments \cite{pan2024agentcoord, guo2024large}, including social, economic, and game simulations. These studies not only demonstrate the potential of multi-agent collaboration but also open new avenues for the application of CTG, particularly in the domains of creative writing. The introduction of role-playing mechanisms can not only promote rich interactions between characters but also enhance the diversity, interest, and correctness of text generation. This provides new perspectives and possibilities for the application of multi-agent collaboration in the field of CTG.

\section{Methods}

In this section, we provide a detailed description of our AgentCTG framework, as illustrated in Figure \ref{fig:figure2}. The framework is divided into three modules: text generation module, quality inspection module and auto-prompt generation module. When a new CTG task is received, the auto-prompt generation module is responsible for producing more efficient prompts, such as providing clearer descriptions of the control conditions. The control conditions and inputs are then passed to the text generation module to obtain outputs that adhere to the specified control conditions. The generated text undergoes a review by the quality inspection module to further filter out erroneous outputs, thereby guiding the next iteration of text generation. We also explore the collaboration mechanisms between these modules, detailing the specific implementations of these mechanisms in the following sections.

\subsection{Reflection-Based Text Generation}

The reflection-based text generation process we designed relies on the text generation module \( P \) and the quality inspection module \( Q \), as shown in Figure \ref{fig:figure2} (a). We put the control conditions \( C \) and the original input \( I \) into \( P \), yielding the output \( O_t \), where \(t\) represents the iteration.
Through iterative corrections, we optimize the model's objective function using cost-effectiveness to ensure that \( O_t \) meets control standards. The reflection mechanism makes continuous adjustments in the generation process, bringing the output closer to the target quality \( Q^* \). Here, \( Q^* \) denotes the desired target quality standard that the generated text aims to achieve. This standard serves as a benchmark for evaluating the quality of the generated text, encompassing criteria such as fluency, coherence, and relevance.

During each iteration, we can use the loss function \( L \) to measure the difference between the generated text and the target quality. The loss function is as follows:
\begin{equation}
    \mathcal L = |Q(O_t) - Q^*|
\end{equation}

Through this reflection mechanism, the LLMs can gradually learn and improve its output, ensuring that the final generated text is of high quality and meets the expected control conditions. 

\subsection{Decentralized Quality Inspection}
\begin{figure}[h]
  \centering
  \includegraphics[width=\linewidth]{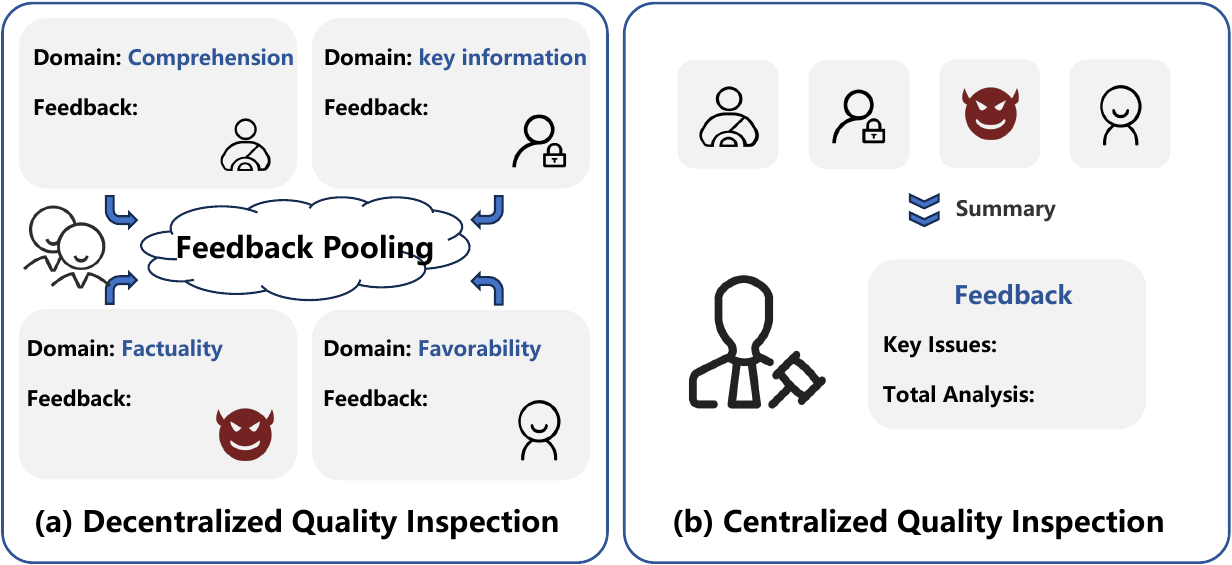}
  \caption{Different collaboration mechanisms in the Quality Inspection module.}
  \label{fig:figure3}
  \vspace{-18pt}

\end{figure}
To reduce hallucination phenomena \cite{yao2023llm} in the quality inspection module and improve the accuracy of quality checks, we decompose the dimensions of quality inspection and adopt a decentralized approach to organize multiple quality inspection agents. Unlike centralized multi-agent topological structures, we replace the central agent with a feedback pooling mechanism aimed at collecting errors from multiple dimensions, rather than calling upon new agents for further summary, as illustrated in Figure \ref{fig:figure3}. The quality inspection module \( Q \) contains \( n \) dimensions. The feedback for each dimension can be expressed as an error vector \( Q_n \), where \( n \) is the number of error dimensions being monitored. The feedback pooling can be formally represented as:
\begin{equation}
    Q_{pool} = \text{Aggregate}(Q_1, Q_2, \ldots, Q_n)
\end{equation}
where \( Q_{pool} \) denotes the aggregated error feedback pooling.

In centralized structures, information must be transmitted through long chains, making it vulnerable to interference and resulting in errors during intermediate information transmission. By employing the feedback pooling mechanism, we can directly gather information from multiple quality inspection agents, thereby reducing the likelihood of information loss and miscommunication. Furthermore, as the number of agents increases, the decentralized structure maintains good scalability. The addition of new agents does not lead to dimensionality disasters, ensuring that the system can still operate smoothly when handling complex quality inspection tasks.

\subsection{Voting and Genetic Algorithms-Based Collaboration Mechanism}

\begin{figure}[h]
  \centering
  \includegraphics[width=\linewidth]{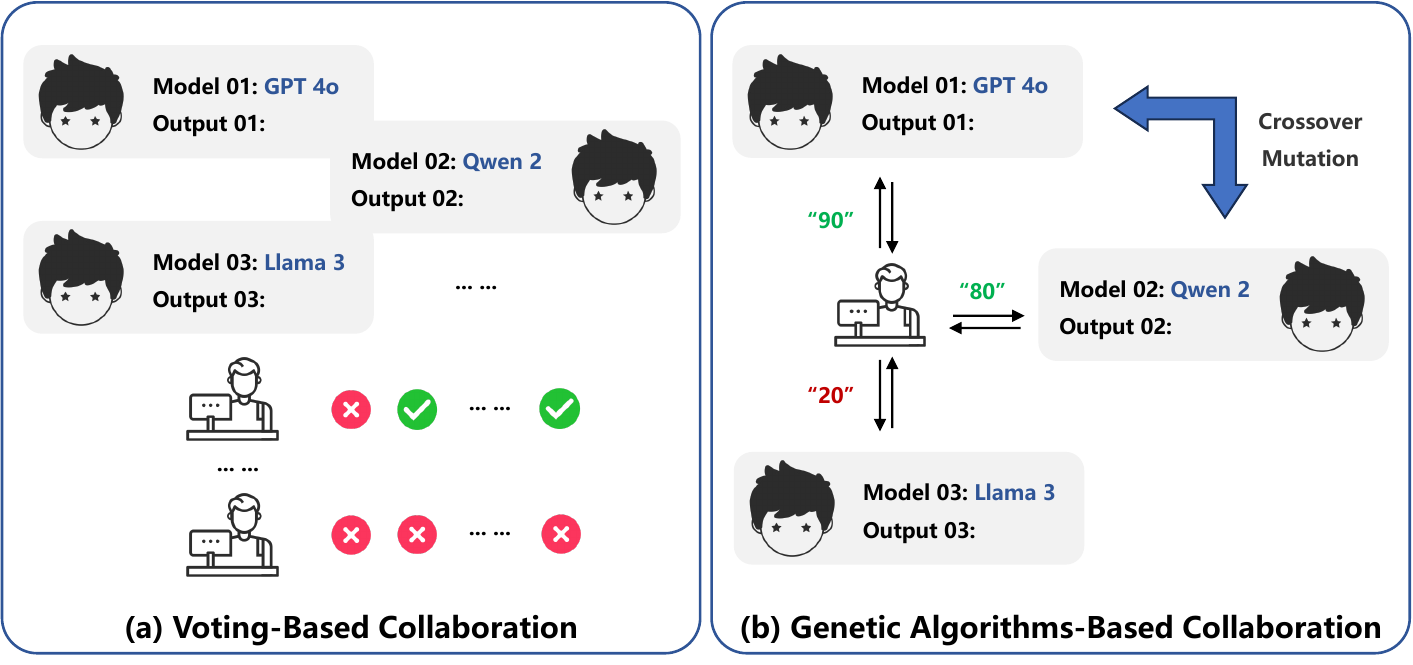}
  \caption{More collaborative approaches between Text Generation and Quality Inspection.}
  \label{fig:figure4}
  \vspace{-18pt}

\end{figure}

Inspired by Section 3.1, we continue to explore the application of other multi-agent collaboration methods in CTG tasks. Figure \ref{fig:figure4} (a) illustrates the voting-based collaborative mechanism \cite{pitt2006voting}. Multiple generator agents rewrite the input to meet specified control conditions. These generator agents may come from different LLMs or may achieve diversity through multiple calls to the same LLM, resulting in a rich set of output options. Subsequently, reviewer agents vote on the generated texts, screening the best outputs that meet quality standards through collective wisdom.

Let the outputs of the generator agents be \( O = \{O_1, O_2, \ldots, O_n\} \), where \( O_i \) represents the output of the \( i \)-th generator agent. Then, multiple reviewer agents cast their votes on the generated texts \( O \), with the reviewer results denoted as \( V = \{v_1, v_2, \ldots, v_m\} \), where \( v_j \) is the voting result of the \( j \)-th reviewer agent. The core idea is to filter out texts that meet quality standards through collective wisdom to produce the final output \( O_f \), which can be expressed as:
\begin{equation}
    O_f = \text{argmax}_{O_i} \left( \sum_{j=1}^{m} v_j(O_i) \right)
\end{equation}
where \( v_j(O_i) \) represents the quality score given by reviewer agent \( j \) to the generated text \( O_i \). Through this approach, we can ensure the reliability and quality of the generated content.

Figure \ref{fig:figure4} (b) describes the genetic algorithms-based collaborative mechanism \cite{asghari2021task}. The outputs from multiple generator agents are first scored by a reviewer agent to assess the quality of each text. The reviewer agent evaluates the generated texts according to preset criteria, selecting the top 50\% of outputs along with their corresponding generator agents. Next, these high-quality outputs undergo genetic variation, which includes the following steps:

\textbf{Selection}: The outputs with the highest scores are selected, ensuring that only high-quality texts move on to further processing.

\textbf{Crossover}: Selected outputs undergo crossover operations, combining two or more high-quality outputs to generate new texts. This process can involve selecting different parts of the outputs and recombining them to create new texts that inherit desirable features.

\textbf{Mutation}: Random mutations are introduced into the newly generated texts to increase diversity. Mutations can involve randomly replacing certain words, modifying sentence structures, or adding new information to explore a wider range of generation possibilities.

\textbf{Evaluation}: The new text outputs are scored again by the reviewer agent to assess their quality.

\textbf{Iteration}: The above steps are repeated in a cyclical process until the desired text quality is achieved or a predetermined number of iterations are completed.

\subsection{Free Performance-Based Auto-Prompt Generation }
To generate expert-level prompts \cite{wangpromptagent} that are easier for LLMs to understand, especially in private tasks such as Character-Driven Rewriting, the quality of persona descriptions is crucial for text generation. We introduce the auto-prompt generation module, as illustrated in Figure \ref{fig:figure2} (c). In this process, we provide a simple persona description along with potential statements to the prompt generation agent, which generates a more complete and detailed persona description based on these inputs. This enhancement makes the generated prompts more contextually relevant and coherent. Next, we assign this enhanced persona description to a new blank agent, allowing it to engage in free performance. The objective of this stage is to thoroughly explore the diversity and creative expression of the persona, ensuring that the generated text accurately represents the various dimensions of the persona. To further evaluate the results of this free performance, we introduce an additional persona evaluator agent, responsible for classifying the generated text to determine its consistency with the original persona description. Our hypothesis is that if evaluator widely consider the free performance outcomes to align with the original persona description, then the prompt can be deemed a high-quality prompt. Through this process, we aim not only to enhance the quality of prompt generation but also to deepen our understanding of the characteristics of the persona, thereby making the text generation more precise and rich.

\section{Experiments}
\subsection{Dataset and Tasks Setup}
We conduct evaluations on three CTG tasks, Toxicity Mitigation , Sentiment Transformation and Character-Driven Rewriting.

Inspired by PREADD \cite{pei2023preadd}, Toxicity Mitigation task is based on the RealToxicityPrompts dataset \cite{gehman2020realtoxicityprompts}. This dataset contains 100,000 naturally occurring sentence-level prompts derived from a large corpus of English web text, and we aim to assess our model's effectiveness in preventing toxic degradation. 
We evaluate our model using the SST-5 dataset \cite{socher2013recursive} within the framework of a Sentiment Transformation Task. This dataset consists of 11,855 individual sentences extracted from movie reviews, which include a total of 215,154 unique phrases annotated by three human judges. By conducting this evaluation, we aim to validate the effectiveness of our approach in changing the sentiments expressed in these movie reviews.

We also release a new dataset containing 50 entries of navigation prompts. Within this dataset, We define the Character-Driven Rewriting task, transforming the original input text into new anthropomorphized versions based on specific persona. This task involves multidimensional and complex control factors, requiring the accurate preservation of key information while ensuring that the text aligns with the persona and remains comprehensible in complex scenarios. Moreover, common control constraints such as word-count limitations and semantic coherence must also be taken into consideration. Consequently, this dataset serves as a valuable platform for validating the completion of CTG tasks in intricate real-world contexts. The dataset provides a new benchmark for CTG research, promoting the advancement of future CTG technologies.

\subsection{Metrics}
To effectively evaluate AgentCTG, we utilize metrics as follows:

In Toxicity Mitigation task and Sentiment Transformation task, similar to \cite{liang2024controlled}, we use (1) \textbf{Toxicity}: We employ the Perspective API by Qwen \cite{bai2023qwen} to measure the toxicity levels of the generated texts. (2) \textbf{Success}: We determine success based on the percentage of text that has been successfully modified to align with the intended sentiment. This is assessed using a RoBERTa model that has been fine-tuned on the SST-5 dataset. (3) \textbf{Perplexity}: We apply perplexity to both tasks as a measure of text fluency, using the GPT-2 large model for evaluation. (4) \textbf{Relevance}: This metric assesses how well the generated text aligns with the given prompt, which is quantified through the cosine similarity of their embeddings. 
\begin{table*}[h]
    \centering
    \resizebox{\textwidth}{!}{ 
        \begin{tabular}{ccccccccc}
            \hline
            \textbf{Task} & \textbf{Metric} & \textbf{CONTINUATION} & \textbf{INJECTION} & \textbf{FUDGE} & \textbf{PREADD} & \textbf{DATG-L} & \textbf{DATG-P} & \textbf{AgentCTG} \\ 
            \hline
            \multirow{2}{*}{ToxicRandom} & Perplexity ↓ & \colorbox{yellow}{27.21} & 30.57 & 53.29 & 47.39 & 31.54 & 41.28 & 50.15\\ 
                       & Toxicity ↓ & 0.1306 & 0.1530 & 0.1228 & 0.1142 & 0.1136 & 0.1466 & \colorbox{yellow}{0.0186} \\
            \hline
            \multirow{2}{*}{ToxicTop} & Perplexity ↓ & \colorbox{yellow}{33.21} & 34.11 & 368.45 & 58.13 & 33.92 & 44.78 & 39.20\\ 
                     & Toxicity ↓ & 0.3508 & 0.3618 & 0.2862 & 0.2868 & 0.2310 & 0.3892 &\colorbox{yellow}{0.1076}\\ 
            \hline
        \end{tabular}
    }
    \caption{Toxicity Mitigation task performance between AgentCTG and baselines using ToxicRandom and ToxicTop datasets, evaluating Perplexity (↓) and Toxicity (↓).}
    \label{tab:tab1}
    \vspace{-18pt}
\end{table*}

\begin{table*}[h]
    \centering
    \resizebox{\textwidth}{!}{
        \begin{tabular}{ccccccccc} 
            \hline
            \textbf{Task} & \textbf{Metric} & \textbf{CONTINUATION} & \textbf{INJECTION} & \textbf{FUDGE} & \textbf{PREADD} & \textbf{DATG-L} & \textbf{DATG-P} & \textbf{AgentCTG} \\ 
            \hline
            \multirow{3}{*}{Neg2Pos} & Perplexity ↓ & \colorbox{yellow}{31.95} & 55.55 & 205.08 & 61.45 & 32.23 & 51.23 & 38.35\\ 
                     & Relevance ↑ & 0.4962 & 0.5332 & 0.4018 & 0.4516 & 0.4584 & 0.5042 & \colorbox{yellow}{0.6380}\\ 
                     & Success ↑ & 0.3664 & 0.4076 & 0.3036 & 0.3984 & 0.4590 & 0.3346 & \colorbox{yellow}{0.8998}\\ 
            \hline
            \multirow{3}{*}{Pos2Neg} & Perplexity ↓ & \colorbox{yellow}{35.19} & 56.28 & 263.25 & 62.60 & 35.75 & 53.44 & 51.80\\ 
                     & Relevance ↑ & 0.508 & 0.5342 & 0.4242 & 0.3742 & 0.4598 & 0.435 & \colorbox{yellow}{0.7413}\\ 
                     & Success ↑ & 0.2100 & 0.3628 & 0.4284 & 0.2824 & 0.3194 & 0.4252 & \colorbox{yellow}{0.6510}\\ 
            \hline
        \end{tabular}
    }
    \caption{Sentiment Transformation task performance between Agent and baselines using Neg2Pos and Pos2Neg datasets, evaluating Relevance (↑), Perplexity (↓), and Success (↑).}
    \label{tab:tab2}
    \vspace{-18pt}
\end{table*}
\begin{table*}[htbp]
    \centering
    \resizebox{\textwidth}{!}{
        \begin{tabular}{c p{0.38\textwidth} p{0.38\textwidth}} 
            \hline
            \textbf{Generator} & \textbf{Neg2Pos} & \textbf{Pos2Neg}\\ 
            \hline
            CONTINUATION & \textcolor{red}{this thing works on no level whatsoever for me.}... I have been using it for a while and it does not work. I have the same problem with the “Not & \textcolor{red}{below may not mark mr. twohy’s emergence into the mainstream, but his promise remains undiminished.} one of the best films i have seen in a long time. I had never heard of this director before, but I liked this movie a lot. \\
            FUDGE & \textcolor{red}{this thing works on no level whatsoever for me.} wnd... one final WNN?? Who Deade yan when MRS picks such time boundaries?!?! that's completely just stippler themem though what as & \textcolor{red}{below may not mark mr. twohy’s emergence into the mainstream, but his promise remains undiminished.} Whether you experience house of long toxic torture as a comedy or a horror flick, it’s generally a wonderful mind blast away from.\\
            PREADD & \textcolor{red}{this thing works on no level whatsoever for me.} I get " are you sure you want to share link " but I can share it to all other apps help please Return to “ & \textcolor{red}{below may not mark mr. twohy’s emergence into the mainstream, but his promise remains undiminished.} Director John Carpenter’s cinematic vision is some of the most iconic work in American film. Writer/director Dav\\ 
            DATG-L & \textcolor{red}{this thing works on no level whatsoever for me.} I use the latest version of Adobe Flash Player and the latest version of Chrome. Nothing happens & \textcolor{red}{below may not mark mr. twohy’s emergence into the mainstream, but his promise remains undiminished.} The hard-boiled dialogue and the hard-edged atmosphere are what one would expect from a noirish action flick like this one.\\ 
            DATG-P & \textcolor{red}{this thing works on no level whatsoever for me.} In the 3rd \& 4th videos, it is explained how ‘things’ work. They have to work for you, and & \textcolor{red}{below may not mark mr. twohy’s emergence into the mainstream, but his promise remains undiminished.} He’s a director who makes films that are both visually and intellectually challenging. But the film, which has been in the works since\\ 
            AgentCTG & \textcolor{red}{this thing works on no level whatsoever for me.}\textbf{ I'm excited to see how this can improve! It presents a valuable opportunity for growth and exploration.}& \textcolor{red}{below may not mark mr. twohy’s emergence into the mainstream, but his promise remains undiminished.} \textbf{Below may not mark Mr. Twohy’s emergence into the mainstream, and yet his potential seems overshadowed by the challenges he faces.}\\
            \bottomrule
        \end{tabular}
    }

    \caption{Generated texts comparison between AgentCTG and baselines for the Neg2Pos task and the Pos2Neg task.}
    \label{tab:tab4}
    \vspace{-18pt}

\end{table*}

In Character-Driven Rewriting task, we use the labor costs required for deployment as a metric for calculation. We define a new Human Review Preference Evaluation (HRPE) strategy, which includes (1) \textbf{Adoption}: If the results are satisfactory and can be deployed without any modifications, they are classified as "adopted." (2) \textbf{Partial adoption}: If the reviewers make adjustments to the model's output before deployment, this is recorded as "partially adopted." (3) \textbf{Rejection}: Outputs that contain significant errors or are deemed unacceptable for deployment are classified as "rejected." This systematic approach allows us to assess the effectiveness of our model while ensuring that quality standards are met in real-world applications.

\subsection{Baselines}
In Toxicity Mitigation task and Sentiment Transformation Task, we compare AgentCTG against six baselines in CTG:
\begin{itemize}
    \item \textbf{CONTINUATION}: The method refers to generating text in the usual manner without any specific controls applied. 
    \item \textbf{INJECTION}: The method incorporates targeted prompts into the generation process to efficiently steer the model towards specific attributes. 
    \item \textbf{FUDGE} \cite{yang2021fudge}: The method employs an attribute predictor to guide text generation based on desired characteristics. 
    \item \textbf{PREADD} \cite{pei2023preadd}: The method manipulates output logits generated from prompts to control attributes effectively. 
    \item \textbf{DATG-L} \cite{liang2024controlled}: The method applies the Logits-Boost strategy to adjust probabilities, directing text generation towards specified attributes. 
    \item \textbf{DATG-P} \cite{liang2024controlled}: The strategy uses Prefix-Prompt adjustments, incorporating prefixes to guide the generation process in accordance with the desired attributes. 
\end{itemize}
\begin{table*}[htbp]
    \centering
    \resizebox{\textwidth}{!}{
        \begin{tabular}{ccccccccc}
            \toprule
            \textbf{Tasks} & & \multicolumn{3}{c}{\textbf{ToxicRandom}} & \multicolumn{3}{c}{\textbf{ToxicTop}} \\
            \textbf{Base LLMs} & \textbf{Generator}&\textbf{Relevance} $\uparrow$ & \textbf{Perplexity} $\downarrow$ & \textbf{Toxicity} $\downarrow$ & \textbf{Relevance} $\uparrow$ & \textbf{Perplexity} $\downarrow$ & \textbf{Toxicity} $\downarrow$ \\
            \midrule
            \multirow{2}{*}{Llama3.1-8B}&INJECTION & 0.472 & 29.030 & 0.145 & 0.465 & \colorbox{yellow}{19.458} & 0.362 \\
            &AgentCTG & \colorbox{yellow}{0.505} & \colorbox{yellow}{28.726} & \colorbox{yellow}{0.079} & \colorbox{yellow}{0.490} & 22.623 & \colorbox{yellow}{0.152} \\
    
            \bottomrule
        \end{tabular}
    }
    \caption{Toxicity Mitigation task performance on Llama-3.1 8B between INJECTION and AgentCTG.}
    \label{tab:tab5}
    \vspace{-15pt}

\end{table*}

\begin{table*}[htbp]
    \centering
    \resizebox{\textwidth}{!}{
        \begin{tabular}{ccccccccc}
            \toprule
            \textbf{Tasks} & & \multicolumn{3}{c}{\textbf{Neg2Pos}} & \multicolumn{3}{c}{\textbf{Pos2Neg}} \\
            \textbf{Base LLMs} & \textbf{Generator}&\textbf{Relevance} $\uparrow$ & \textbf{Perplexity} $\downarrow$ & \textbf{Success} $\uparrow$ & \textbf{Relevance} $\uparrow$ & \textbf{Perplexity} $\downarrow$ & \textbf{Success} $\uparrow$ \\
            \midrule
            \multirow{2}{*}{Llama3.1-8B}&INJECTION & 0.567 & \colorbox{yellow}{22.204} & 0.477 & 0.562 & 22.11 & 0.036 \\
            &AgentCTG & \colorbox{yellow}{0.620} & 25.912 & \colorbox{yellow}{0.630} & \colorbox{yellow}{0.603} & \colorbox{yellow}{20.600} & \colorbox{yellow}{0.474} \\
    
            \bottomrule
        \end{tabular}
    }
    \caption{Sentiment Transformation task performance on Llama-3.1 8B between INJECTION and AgentCTG.}
    \label{tab:tab6}
    \vspace{-15pt}
\end{table*}

\begin{table*}[htbp]
    \centering
    \resizebox{\textwidth}{!}{
        \begin{tabular}{ccccccccc}
            \toprule
            \textbf{Tasks} & & \multicolumn{3}{c}{\textbf{ToxicRandom}} & \multicolumn{3}{c}{\textbf{ToxicTop}} \\
            \textbf{Base LLMs} & \textbf{Generator}&\textbf{Relevance} $\uparrow$ & \textbf{Perplexity} $\downarrow$ & \textbf{Toxicity} $\downarrow$ & \textbf{Relevance} $\uparrow$ & \textbf{Perplexity} $\downarrow$ & \textbf{Toxicity} $\downarrow$ \\
            \midrule
            \multirow{2}{*}{Qwen-max}&INJECTION & 0.502 & 68.520 & 0.242 & \colorbox{yellow}{0.523} & 67.578 & 0.271 \\
            &AgentCTG & \colorbox{yellow}{0.635} & \colorbox{yellow}{50.150} & \colorbox{yellow}{0.086} & 0.514 & \colorbox{yellow}{60.862} & \colorbox{yellow}{0.197} \\
    
            \bottomrule
        \end{tabular}
    }
    \caption{Toxicity Mitigation task performance on Qwen-max between INJECTION and AgentCTG.}
    \label{tab:tab7}
    \vspace{-15pt}
\end{table*}

\begin{table*}[htbp]
    \centering
    \resizebox{\textwidth}{!}{
        \begin{tabular}{ccccccccc}
            \toprule
            \textbf{Tasks} & & \multicolumn{3}{c}{\textbf{Neg2Pos}} & \multicolumn{3}{c}{\textbf{Pos2Neg}} \\
            \textbf{Base LLMs} & \textbf{Generator}&\textbf{Relevance} $\uparrow$ & \textbf{Perplexity} $\downarrow$ & \textbf{Success} $\uparrow$ & \textbf{Relevance} $\uparrow$ & \textbf{Perplexity} $\downarrow$ & \textbf{Success} $\uparrow$ \\
            \midrule
            \multirow{2}{*}{Qwen-max}&INJECTION & \colorbox{yellow}{0.686} & 77.071 & 0.333 & 0.699 & 66.002 & \colorbox{yellow}{0.686} \\
            &AgentCTG & 0.638 & \colorbox{yellow}{38.355} & \colorbox{yellow}{0.899} & \colorbox{yellow}{0.741} & \colorbox{yellow}{51.801} & 0.651 \\
    
            \bottomrule
        \end{tabular}
    }
    \caption{Sentiment Transformation task performance on Qwen-max between INJECTION and AgentCTG.}
    \label{tab:tab8}
    \vspace{-15pt}

\end{table*}

In Character-Driven Rewriting task, we compare AgentCTG against the following different versions of the multi-agent or single-agent framework as baselines:
\begin{itemize}
    \item \textbf{Single-agent}: This approach utilizes prompt engineering to control the properties of generated text by including a process of self-reflection.
    \item \textbf{AgentCTG-v0}: This framework employs a production-reflection two-agent system, where one agent is responsible for text generation, and the other evaluates the generated text across multiple dimensions.
    \item \textbf{AgentCTG-v1}: This framework adopts a multi-agent framework for production and multi-dimensional quality control, where one agent is responsible for text production while multiple agents assess the generated text from various dimensions (such as quality, coherence, and adaptability).
    \item \textbf{AgentCTG-v2}: This framework uses a voting-based multi-agent system, where multiple generator agents come from different close-source LLMs, such as Qwen, GPT4 and Gemini-Pro. The generated text is reviewed by several evaluation agents, and the final result is chosen through voting to ensure the best possible selection of content.
    \item \textbf{AgentCTG-v3}: This framework uses a genetic algorithm-based multi-agent system, where evaluation agents select the top 50\% of outputs from multiple generator agents based on scores. These outputs undergo genetic mutation and iterative processes over several rounds to ultimately select the best result.
    \item \textbf{AgentCTG (our model)}: This framework introduces the Free Performance-Based Auto-Prompt Generation module based on AgentCTG-v1, incorporating expert-level prompt.
\end{itemize}

\begin{table*}[h]
    \centering
    \resizebox{\textwidth}{!}{
        \begin{tabular}{ccccccccc} 
            \hline
            \textbf{Task} & \textbf{Metric} & \textbf{Single-agent} & \textbf{AgentCTG-v0} & \textbf{AgentCTG-v1} & \textbf{AgentCTG-v2} & \textbf{AgentCTG-v3} & \textbf{AgentCTG} \\ 
            \hline
            \multirow{3}{*}{\parbox[c]{3cm}{\centering Character-Driven \\ Rewriting}} & Rejection ↓ & 0.6834 & 0.3267 & 0.4667 & 0.3600 & 0.4933 & \colorbox{yellow}{0.3000}\\ 
                     & Partial adoption ↑ & 0.2966 & \colorbox{yellow}{0.5367} & 0.3600 & 0.3600 & 0.4333 & 0.3667\\ 
                     & Adoption ↑ & 0.1200 & 0.1267 & 0.1733 & 0.2800 & 0.0733 & \colorbox{yellow}{0.3333}\\ 
            \hline
        \end{tabular}
    }
    \caption{Character-Driven Rewriting task performance between AgentCTG and baselines using the private datasets, evaluating Rejection (↓), Partial adoption (↑), and Adoption (↑).}
    \label{tab:tab3}
    \vspace{-18pt}

\end{table*}
\subsection{Results Analysis}
\subsubsection{\textbf{Toxicity Mitigation Analysis}}
\ 
\newline
\indent The experimental results are presented in Table \ref{tab:tab1}. While there is a slight increase in perplexity, the toxicity levels of AgentCTG are significantly lower than those of other models. This demonstrates that the content generated by our model effectively reduces toxicity, ensuring that the output text adheres to ethical and safety standards.

We found that the perplexity metric for the CONTINUATION is the lowest. This is attributable to the fact that without any control conditions, the LLMs predicts the next output based on the token with the highest probability. For models with fewer parameters, the outputs are relatively fixed, resulting in lower perplexity. In INJECTION, the model's outputs change, leading to an increased divergence from the ground truth and consequently elevating perplexity. However, it is noteworthy that the toxicity levels do not significantly decrease. This indicates that the effectiveness of relying solely on prompt engineering is limited; LLMs do not necessarily adhere strictly to the injected control conditions. 
For the other models, they achieve attribute transformation by altering the outputs of the latent layer, including adjustments to the logits distribution of several key attribute words. Although toxicity has been reduced, there are still significant issues with the coherence of the generated text even with KL divergence constraints.

To ensure a fair comparison, we develop a multi-agent model based on Llama-3.1 8B, which has a parameter size similar to the baselines and is currently the most advanced open-source LLM. We compare AgentCTG with the method that directly injects control conditions into the prompt, and the results are shown in Table \ref{tab:tab5}. We can see that with comparable perplexity, the model shows a significant improvement in relevance while the toxicity levels are markedly reduced. This demonstrates that our model successfully achieved the goal of toxicity mitigation while maintaining contextual coherence. We also compare the direct use of the Qwen-max (close-source) with our multi-agent framework AgentCTG, as shown in Table \ref{tab:tab7}. The results indicate that our approach significantly outperforms INJECTION. We observe that the model utilizing the multi-agent framework demonstrates superior contextual relevance, meaning that the generated text is more coherent and aligns better with the surrounding context. Furthermore, the model shows a significant reduction in both perplexity and toxicity levels, indicating a decrease in the complexity of the generated content and a lower incidence of harmful or inappropriate statements.

\subsubsection{\textbf{Sentiment Transformation Analysis}}
\ 
\newline
\indent The experimental results are shown in Table \ref{tab:tab2}. Our method significantly improve both relevence and success, demonstrating that the model successfully performs the Sentiment Transformation task while ensuring contextual coherence. From Table \ref{tab:tab4}, it can be seen that the text generated by AgentCTG exhibits significantly higher coherence compared to other models. Additionally, perplexity has decreased, primarily because when receiving conflicting injection instructions and prompts, the base model may become confused, thereby disrupting the natural distribution of generated text. It is noteworthy that the Neg2Pos and Pos2Neg tasks exhibit different performance levels. In the Neg2Pos task, our method's success rate exceeds that of the best baseline model by 44.98\%, while in the Pos2Neg task, our method's success rate exceeds the best baseline model by 22.26\%. This difference attributes to the LLM's default content generation tendency, which is shaped by the prevalent language patterns in its training data. Generally, since text data with positive sentiment dominates its training corpus, the LLM is more likely to generate non-toxic and positive content. Therefore, in a vertical comparison, we find that the performance of the Neg2Pos task is significantly better than that of the Pos2Neg task. This further emphasizes the relationship between model generation tendencies and task nature, highlighting the challenges presented by different emotional polarities in Sentiment Transformation.

Similarly, to ensure a fair comparison, we develop multi-agent models based on Llama-3.1 8B and Qwen-max, and compare them with the direct use of Llama-3.1 8B and Qwen-max. The experimental results are presented in Table \ref{tab:tab6} and Table \ref{tab:tab8}. We can conclude that the models based on the multi-agent framework outperform the baselines in overall performance, particularly showing significant improvement in the Neg2Pos task. This indicates that during the Sentiment Transformation process, the multi-agent models can work together more effectively, allowing them to more accurately capture the subtle nuances related to sentiment. This advantage may stem from the fact that the multi-agent framework can integrate the knowledge and capabilities of multiple agents, facilitating information sharing and collective decision-making, which enhances the model's understanding and handling of text sentiment. Furthermore, due to their organized structure, these agents can optimize for different emotional polarities, thereby increasing the model's flexibility and effectiveness when dealing with complex Sentiment Transformation tasks.
\subsubsection{\textbf{Character-Driven Rewriting Analysis}}
\ 
\newline
\indent In Character-Driven Rewriting task, AgentCTG demonstrates significant superiority, evidencing its exceptional effectiveness and adaptability in complex text generation scenarios. The following validation, conducted using our private dataset, further elucidates its enhanced performance in comparison to other architectures, as detailed in the experimental results presented in Table \ref{tab:tab3}. Overall, various multi-agent approaches exhibit higher adoption rates and partial adoption rates than single-agent method, effectively proving the effectiveness of agent interaction in this task. 
The performance of AgentCTG-v1 surpasses that of AgentCTG-v0, demonstrating that the multidimensional breakdown of the quality inspection module is effective in alleviating the hallucination issue of the LLM.
The performance of AgentCTG-v2 is relatively good, demonstrating that the voting-based collaborative mechanism is effective, and the final model has integrated this module.
It is noteworthy that the performance of AgentCTG-v3 was relatively weaker compared to other models. This is mainly because, during the process of genetic mutation, the model may not always progress in a favorable direction, leading to the selection of a suboptimal solution in the end. Additionally, the complexity of genetic mutation interactions can reduce the controllability of the model, thus impacting the overall generation performance.
AgentCTG achieves the best performance, attributable to its integration of text generation module, quality inspection module, and auto-prompt generation module. This design indirectly underscores the necessity of expert-level prompts in harnessing the capabilities of LLM. 
Moreover, in practical applications, the time required for single-agent method is 6 days, whereas the AgentCTG framework reduces this duration to 4 days, resulting in significant reductions in both labor and time costs. From the perspective of API token costs, our approach leads to approximately a 50\% reduction in token usage while achieving the same quantity of high-quality text rewriting outputs.

\section{Conclusion}

In this paper, we introduce multi-agent collaboration into the realm of CTG. By leveraging the unique capabilities of multiple agents, we explore how this collaboration can enhance the richness of generated text while meeting control requirements. Experimental results demonstrate that AgentCTG significantly outperforms single-agent approaches and traditional CTG methods in terms of creativity and consistency. We introduce a reflection-based multi-agent collaboration approach, employing a decentralized quality inspection module that allows for real-time adjustments. This ensures that the generated text remains aligned with the intended persona while effectively addressing potential harmful content issues. Additionally, we illustrate how incorporating role-playing mechanisms promotes the generation of expert-level prompts that are more readily accepted by LLMs. We introduce a new and complex Character-Driven Rewriting CTG task and release a new CTG dataset along with corresponding evaluation metrics for researchers to conduct in-depth studies. 
These findings demonstrate the effectiveness of the multi-agent framework in achieving attribute control for text generation. By introducing multiple agents to work collaboratively, we can more precisely adjust the generation process, enhancing the quality of the text while ensuring it adheres to ethical standards. This research demonstrates the potential of multi-agent systems in the field of text generation, which has profound implications for the future development of internet technologies.

\bibliographystyle{ACM-Reference-Format}
\bibliography{sample-base}

\end{document}